\documentclass[letterpaper,10pt,conference]{ieeeconf}

\IEEEoverridecommandlockouts                              % This command is only needed if 
\usepackage{graphics} % for pdf, bitmapped graphics files
\usepackage{epsfig} % for postscript graphics files
\usepackage{mathptmx} % assumes new font selection scheme installed
\usepackage{times} % assumes new font selection scheme installed
\usepackage{amsmath} % assumes amsmath package installed
\usepackage{amssymb}  % assumes amsmath package installed
\usepackage{pifont}

\usepackage{booktabs}
\usepackage{xcolor}
\title{TacOT: Learning Contact-Rich Dexterous Manipulation from\\
Human Demonstrations via Tactile-Guided Optimal Transport}

\author{
{Xingting Li}$^{1*}$,
{Yifan Han}$^{2*}$,
{Zijian Lin}$^{1*}$,
{Wei Hou}$^{1}$,
{Chuqiao Lyu}$^{3}$,
{Shoujie Li}$^{4\dagger}$,
{Wenbo Ding}$^{1\dagger}$
\thanks{{$*$}These authors contributed equally to this work.}
\thanks{{$^1$}Shenzhen Ubiquitous Data Enabling Key Lab, Shenzhen International Graduate School, Tsinghua University, Shenzhen 518055, China.}
\thanks{{$^2$}School of Artificial Intelligence, Shanghai Jiao Tong University, Shanghai 200030, China.}
\thanks{{$^3$}Xspark AI, Shenzhen, China.}
\thanks{{$^4$}School of Mechanical and Aerospace Engineering, Nanyang Technological University, Singapore 639956, Singapore.}
\thanks{{$\dagger$}\raggedright
Corresponding authors: Shoujie Li, lsj20@mails.tsinghua.edu.cn;
Wenbo Ding, ding.wenbo@sz.tsinghua.edu.cn.}
}

\begin{document}

\maketitle
\thispagestyle{empty}
\pagestyle{empty}

%%%%%%%%%%%%%%%%%%%%%%%%%%%%%%%%%%%%%%%%%%%%%%%%%%%%%%%%%%%%%%%%%%%%%%%%%%%%%%%%
\begin{abstract}

Learning contact-rich dexterous manipulation from human demonstrations
provides a scalable source of interaction data, yet transferring such
skills to robots remains challenging due to unreliable human--robot
correspondence. Existing human-to-robot transfer methods typically
rely on visual appearance or motion similarity, which may associate
similar motions with different contact states and force patterns.
Tactile dynamics provide interaction-aware cues to distinguish
manipulation processes with similar motions but different contact states.
We introduce tactile-guided optimal transport (TacOT),
a framework for human-to-robot contact-rich manipulation. TacOT leverages action--tactile dynamic
time warping to identify human--robot demonstration correspondences with
consistent interaction dynamics and uses these correspondences to guide
soft optimal transport alignment in a shared policy representation space.
This enables human demonstrations to provide contact-rich supervision for
robot policy learning without requiring predefined frame-level
human--robot pairing.
Across four real-world dexterous manipulation tasks, TacOT improves
closed-loop success rates over action-guided OT by up to 17 points on in-distribution tasks and 20 points under targeted human-to-robot out-of-distribution transfer. Further analyses show that
tactile-guided correspondence selects demonstration pairs with more
consistent contact dynamics and produces latent representations that
better reflect interaction-state evolution.
These results demonstrate that tactile dynamics provide an effective
semantic signal for establishing reliable human-to-robot correspondence
in contact-rich dexterous manipulation.

\end{abstract}

%%%%%%%%%%%%%%%%%%%%%%%%%%%%%%%%%%%%%%%%%%%%%%%%%%%%%%%%%%%%%%%%%%%%%%%%%%%%%%%%
\section{INTRODUCTION}

Learning contact-rich skills with dexterous robot hands remains
constrained by the scarcity of robot demonstrations. Collecting
multi-finger interactions on physical robots is slow, labor-intensive,
and difficult to scale across objects, scenes, and contact conditions.
Automated demonstration generation and visuo-tactile pretraining
improve data efficiency~\cite{dexmimicgen2025,vital2025}, while large
multi-embodiment datasets broaden the coverage of robot
learning~\cite{open_x_embodiment}. Yet dexterous manipulation remains
underrepresented because its critical variations lie not only in
appearance and motion, but also in how contact is established,
maintained, and released.

Human demonstrations offer a scalable source of this missing
interaction diversity. Wearable and in-the-wild systems can capture
natural hand motion, visual observations, and contact signals across
diverse manipulation behaviors~\cite{umi,mimicplay,dexwild2025,mimictouch}.
However, leveraging human data raises a fundamental correspondence
question: which human interactions provide meaningful supervision
for robot behavior despite differences in embodiment?
Existing approaches commonly rely on visual similarity or
end-effector motion~\cite{egomimic,egobridge}. These cues are
insufficient for contact-rich dexterous manipulation: similar wrist
motions can correspond to free-space motion, contact establishment,
or forceful interaction, while morphologically different human and
robot motions may realize the same physical behavior.
Consequently, demonstrations can appear aligned in vision or
kinematics while differing substantially in finger contacts
and force distributions.
\begin{figure}[t]
    \centering
    \includegraphics[width=0.98\columnwidth]{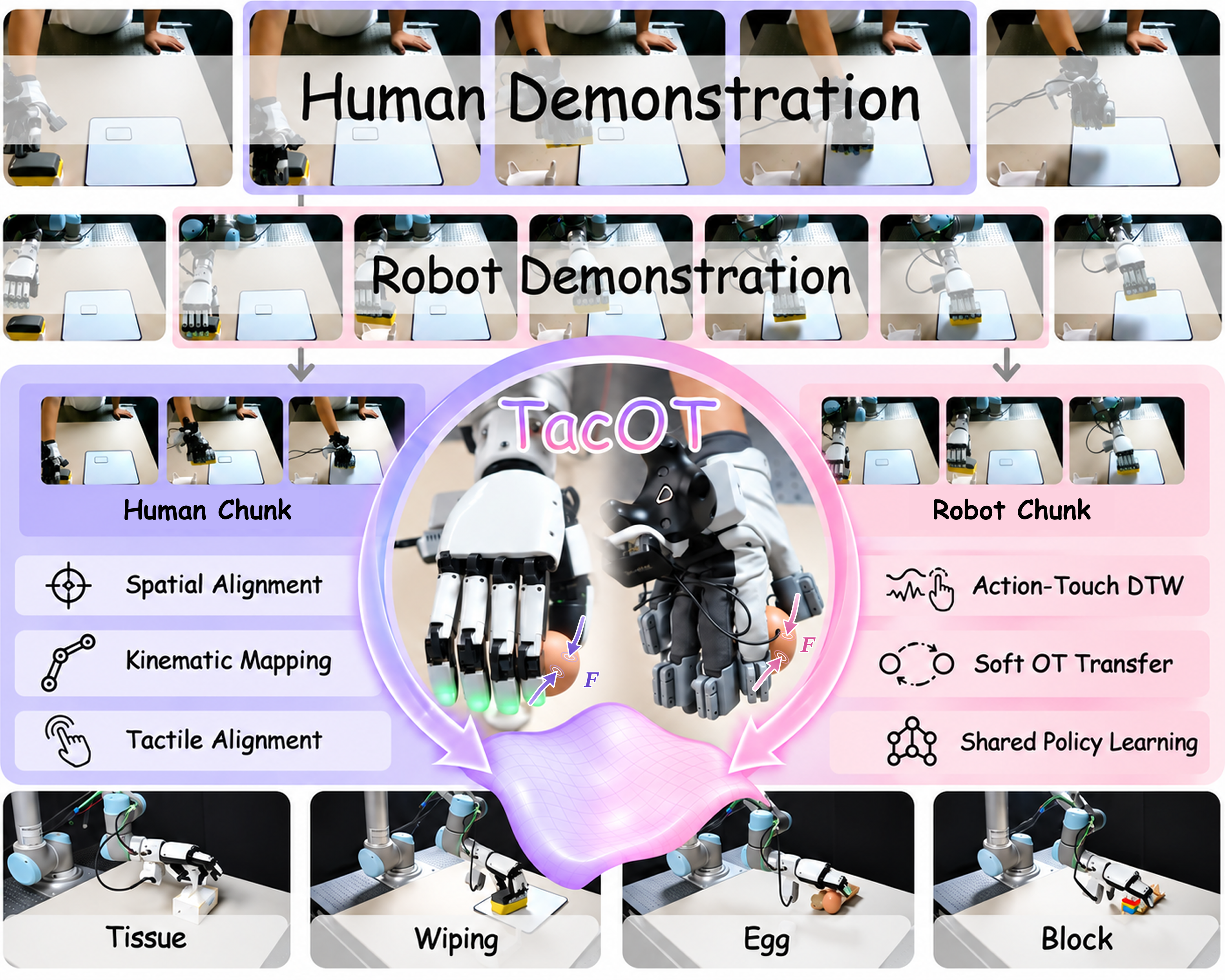}
    \caption{
    Learning contact-rich dexterous manipulation from human demonstrations.
    TacOT uses action--tactile dynamics to establish interaction-consistent
    human--robot correspondence and guide soft optimal transport alignment.
    }
    \label{fig:teaser}
\end{figure}
We therefore cast human-to-robot imitation as identifying
interaction-consistent behavior rather than merely closing
appearance- or motion-domain gaps. The relevant unit of
correspondence is a temporally extended action--contact pattern,
not an individual visually similar frame. Tactile trajectories
disambiguate interaction phases that vision and kinematics leave
uncertain. Here, tactile information serves not only as a policy
observation but also as a supervision signal for matching human
and robot demonstrations with consistent interaction dynamics.
Although prior work has demonstrated the value of touch for
policy conditioning, tactile representation learning, and
cross-sensor transfer~\cite{mimictouch,sparsh,xu2026vt,
transferable_tactile_transformers,tactalign,unitachand},
using joint action--tactile dynamics to establish human–robot behavioral correspondence remains underexplored.

The same interaction may exhibit temporal variations between human and robot demonstrations, and multiple human segments may correspond to similar robot behaviors. Therefore, effective correspondence requires flexible alignment beyond frame-wise matching. Dynamic time warping (DTW) addresses the temporal requirement through order-preserving sequence alignment~\cite{sakoe1978dtw}. To support soft correspondence, we adopt entropy-regularized optimal transport (OT), which couples human and robot policy representations without predefined one-to-one pairing~\cite{cuturi2013sinkhorn,peyre2019cot}. The key challenge is to define a transport cost that reflects interaction compatibility between human and robot behaviors. We use joint action--tactile dynamics to guide this cost, favoring correspondences consistent in both motion and contact.

We present tactile-guided optimal transport (\textbf{TacOT}),
a framework for contact-rich human-to-robot manipulation.
TacOT combines DTW distances computed separately over action
and tactile trajectories to identify interaction-consistent
human--robot demonstration pairs. These pairs guide OT by
reducing their latent transport costs, favoring compatible
interactions while retaining soft coupling across all pairs.
Optimized jointly with a multimodal flow-matching objective,
this alignment allows contact-rich human demonstrations to
supervise a shared policy space anchored by robot data.
Our contributions are as follows:
\begin{itemize}
    \item We formulate human-to-robot imitation for contact-rich
    dexterous manipulation as an interaction-consistent correspondence
    learning problem, using joint action--tactile dynamics to guide
    soft optimal transport alignment.

    \item We develop a matched-sensing human--robot platform and
    a multimodal flow-matching policy for joint learning from
    visual, motion, and tactile demonstration data from human and robot systems.

    \item We evaluate TacOT on four real-world dexterous manipulation
    tasks. Compared with action-guided OT, TacOT improves success rates by up to 17 percentage points in-distribution and 20 percentage points under targeted human-to-robot transfer.
\end{itemize}

%%%%%%%%%%%%%%%%%%%%%%%%%%%%%%%%%%%%%%%%%%%%%%%%%%%%%%%%%%%%%%%%%%%%%%%%%%%%%%%%
\section{RELATED WORK}

\textbf{Dexterous Imitation Learning.}
Dexterous imitation learning models high-dimensional, contact-rich hand--object interactions across tasks and embodiments. Video-based methods derive robot-executable supervision through physics-aware policy learning or hand--object reconstruction and retargeting~\cite{chen2025vividex,paliwal2026dexterous}. Being-H0, VITRA, and Being-H0.5 scale dexterous VLA pretraining through hand-motion modeling and action representations shared across human and robot embodiments~\cite{luo2025being,li2025scalable,luo2026being}, while DexHiL incorporates coordinated arm--hand interventions for policy adaptation~\cite{han2026dexhil}. These methods broaden supervision beyond robot teleoperation, but transferring human manipulation skills still requires alignment between human and robot representations and often robot-specific adaptation data.

\textbf{Tactile Representation and Transfer.}
Recent work learns tactile representations transferable across sensors, objects, and tasks. Self-supervised methods extract reusable tactile features~\cite{sparsh}, while multi-sensor approaches learn representations across heterogeneous visuo-tactile platforms~\cite{anytouch,transferable_tactile_transformers}. MimicTouch incorporates human tactile demonstrations into contact-rich robot learning~\cite{mimictouch}, and UniTacHand learns shared representations for human-to-robot tactile transfer~\cite{unitachand}. These studies emphasize tactile representation learning and transfer; TacOT additionally uses tactile trajectories to guide human--robot demonstration matching during policy learning.

\textbf{Learning from Human Demonstrations.}
MimicPlay learns long-horizon robot behaviors from human play data~\cite{mimicplay}, while EgoMimic co-trains human and robot demonstrations through visual and action-space alignment~\cite{egomimic}. EgoBridge constructs pseudo-pairs from end-effector trajectories to guide optimal-transport alignment between human and robot representations for robots with two-finger grippers~\cite{egobridge}. However, similar motions can involve different contact states in dexterous manipulation. TacOT therefore combines end-effector motion, hand actions, and tactile evolution to identify correspondences consistent in both motion and contact dynamics.

%%%%%%%%%%%%%%%%%%%%%%%%%%%%%%%%%%%%%%%%%%%%%%%%%%%%%%%%%%%%%%%%%%%%%%%%%%%%%%%%
\section{SYSTEM AND METHOD}

\begin{figure}[t]
    \centering
    \includegraphics[width=\columnwidth]{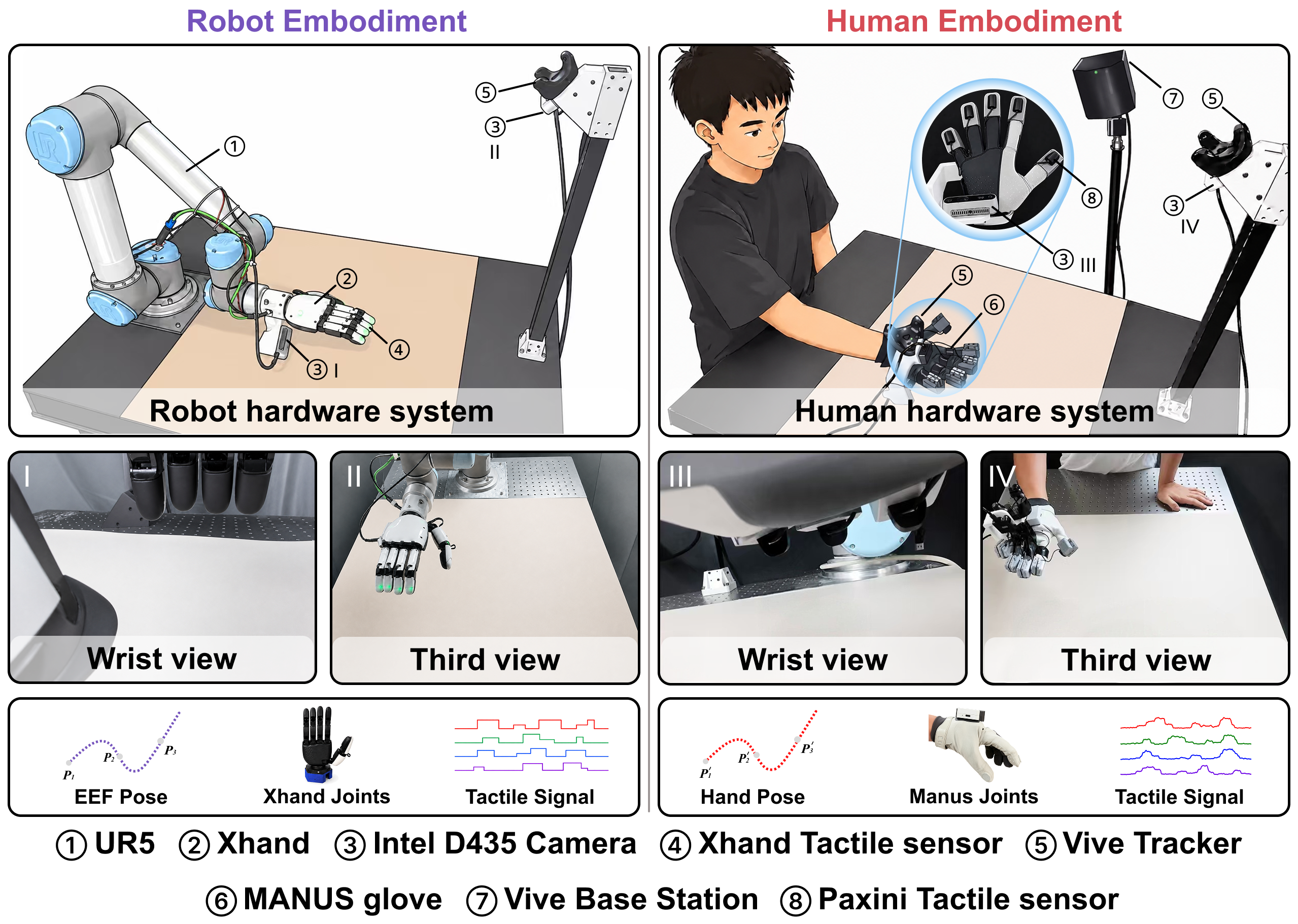}
    \caption{Hardware systems for multimodal demonstration collection.
    The robot and human embodiments provide spatial motion, hand
    configuration, fingertip contact forces, and wrist-view and
    third-person RGB observations.}
    \label{fig:hardware_system}
\end{figure}

\begin{figure*}[t]
    \centering
    \includegraphics[width=0.98\textwidth]{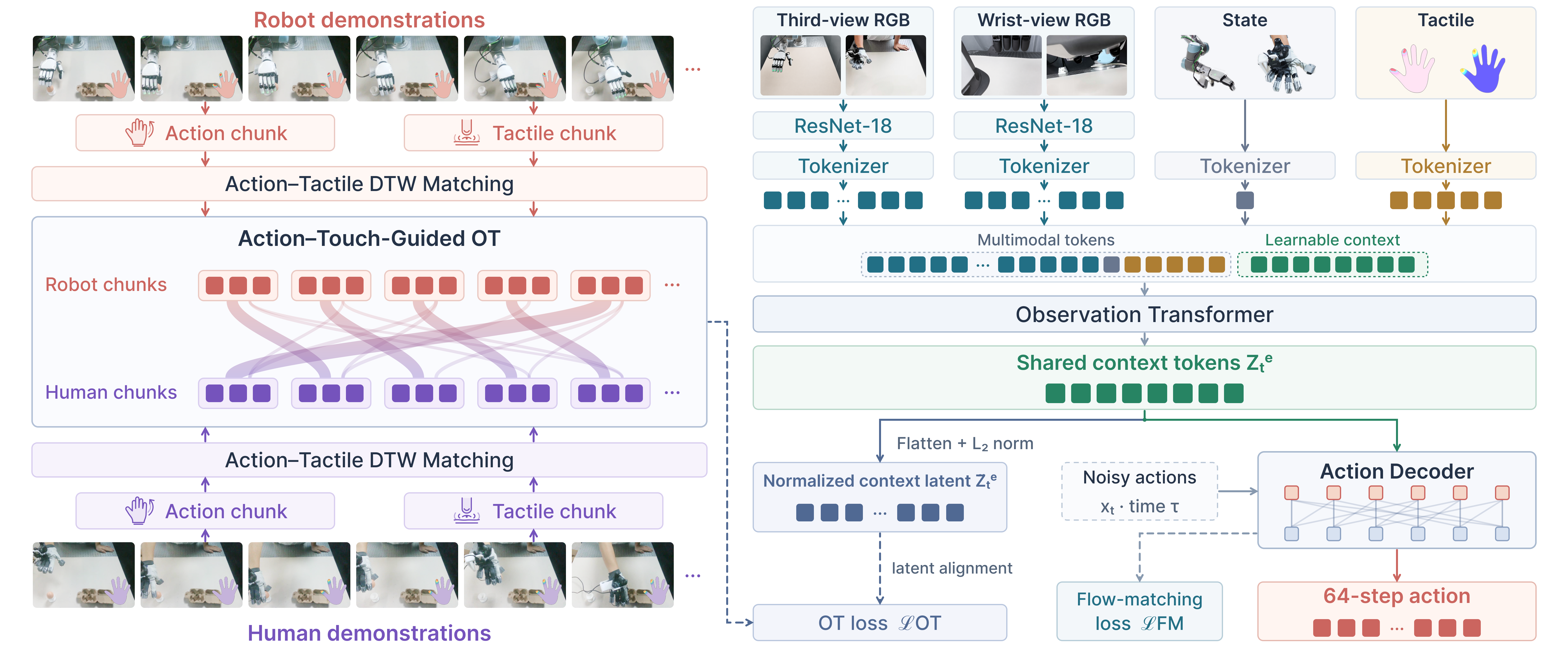}
    \caption{Overview of TacOT. TacOT identifies interaction-consistent human--robot correspondence using
    action--tactile DTW and leverages the correspondence to guide soft optimal
    transport alignment.
    The resulting interaction-consistent alignment enables human demonstrations
    to supervise robot policy learning in a shared representation space.
    }
    \label{fig:pipeline}
\end{figure*}

TacOT learns human-to-robot dexterous manipulation by jointly
leveraging human and robot demonstrations with synchronized visual,
motion, and tactile observations. We first construct matched human and
robot demonstration systems for multimodal data collection
(Fig.~\ref{fig:hardware_system}). The collected observations are then
converted into unified representations for multimodal policy learning
and action--tactile-guided latent alignment, as illustrated in
Fig.~\ref{fig:pipeline}.
The following sections describe the hardware design and data collection
(Sec.~III-A), human–robot representation alignment
(Sec.~III-B), multimodal flow policy learning (Sec.~III-C), and
action--tactile-guided latent alignment (Sec.~III-D).

\subsection{Hardware Design and Data Collection}

To support human-to-robot transfer, we construct two matched-sensing
demonstration systems for multimodal data collection.
The human demonstration system records natural hand--object
interactions, whereas the robot demonstration system maps human motion
to executable arm--hand commands through teleoperation. Both systems
collect visual, motion, and tactile observations for human–robot
representation learning. The human and robot datasets are collected
independently and are not linked by shared timestamps, episode
identifiers, or recorded cross-embodiment pairings.
Fig.~\ref{fig:hardware_system}
summarizes their hardware configurations and sensing modalities.

\textbf{Human demonstration system.}
The human system records hand motion and fingertip contact forces using
a Manus data glove, a Vive Tracker with base stations, and five
three-axis Paxini force sensors. Each sensor is fixed to a Manus
fingertip using a 3D-printed mount, providing repeatable placement
across demonstrations. Two Intel RealSense
D435 cameras record synchronized RGB observations at a resolution of
\(640\times480\), providing wrist-view and third-person views of the
interaction.

\textbf{Robot demonstration system.}
The robot system consists of a UR5 manipulator equipped with an XHand
dexterous hand. During teleoperation, the human wrist trajectory is
mapped to the UR5 end effector (EEF), while glove signals are converted
online into XHand joint commands. The glove and tracker signals used
for teleoperation are not stored and are not used for training. The
robot system records only robot-side EEF and hand trajectories,
fingertip contact forces measured by tactile sensors integrated into the
XHand hand, and dual-view RGB observations using two D435 cameras.

\subsection{Human-Robot Representation Alignment}
\label{sec:cross_embodiment_conversion}

Although both human and robot systems collect visual, motion, and
tactile observations, their raw representations differ due to
embodiment-specific kinematics. We therefore convert human and robot
demonstrations into unified visual, motion, and tactile representations
for downstream learning. Each system records synchronized multimodal
observations at \(30~\mathrm{Hz}\), including wrist-view and third-person
RGB images.

\textbf{Spatial trajectory conversion.}
To establish a common spatial reference across embodiments, human wrist
and XHand poses are expressed in the third-person camera frame \(C\) as
\(\mathbf{T}^{\mathrm{H}}_{C \leftarrow H}\) and
\(\mathbf{T}^{\mathrm{R}}_{C \leftarrow X}\), respectively.
Here, \(H\) and \(X\) denote the human wrist and canonical XHand frames.

\textbf{Joint and action conversion.}
To bridge the morphological differences between the human hand and
XHand, we select 12 semantically corresponding Manus joints, map their
angular ranges to those of the XHand, and arrange them according to the
XHand joint order. This yields a unified 12-DoF hand representation.
End-effector actions are represented as camera-frame position offsets
and wrapped roll--pitch--yaw increments relative to the current pose,
while hand actions are represented as absolute joint angles in the
XHand joint space.
Each target action chunk contains \(K=64\) future steps. Robot chunks
use 64 recorded frames, whereas human chunks are linearly resampled from
32 frames to 64 steps. The same interpolation weights are applied to
action and tactile sequences to preserve their temporal correspondence.

\textbf{Tactile conversion.}
Both systems measure three-axis forces at five fingertips, yielding
15 force channels. Sensor-frame forces are transformed into a shared
canonical frame and arranged in a consistent finger and axis order.
The action and tactile signals are independently normalized to reduce
residual scale differences between embodiments.

\subsection{Multimodal Policy Learning via Flow Matching}
\label{sec:tacot_policy}

TacOT jointly learns from human and robot demonstrations to generate
action chunks conditioned on current multimodal observations.
The policy combines a multimodal observation encoder with a
flow-based action decoder, as illustrated in Fig.~\ref{fig:pipeline}.

For embodiment \(e\in\{H,R\}\), denoting human and robot, respectively,
the observation at timestep \(t\) is
\begin{equation}
    o_t^e = (I_t^e,S_t^e,F_t^e),
    \label{eq:multimodal_observation}
\end{equation}
where \(I_t^e\) contains dual-view RGB images,
\(S_t^e\) comprises the camera-frame pose and 12-DoF hand state,
and \(F_t^e\) contains current 15-D fingertip forces.

Modality-specific encoders convert these observations into tokens.
We prepend \(M=8\) learnable context tokens and process the complete
sequence with a multimodal observation Transformer.
Denoting this encoding process by \(F_\phi\), we obtain
\begin{equation}
    E_t^e = F_\phi(o_t^e),
    \qquad
    Z_t^e = E_t^e[1{:}M]
    \in \mathbb{R}^{M\times d},
    \label{eq:tacot_encoder}
\end{equation}
where \(d\) is the token dimension.
The encoded context tokens \(Z_t^e\) serve as cross-attention memory
for the action Transformer and as representations for
human--robot alignment.

We train the action Transformer using conditional flow matching
to generate action chunks
\(\mathbf{A}_t^e=(a_t^e,\ldots,a_{t+K-1}^e)\), with \(K=64\).
Given a demonstrated action chunk, we sample flow time
\(\tau\sim\mathcal{U}[0,1]\) and Gaussian noise
\(\boldsymbol{\epsilon}\sim\mathcal{N}(\mathbf{0},\mathbf{I})\),
and construct
\begin{equation}
    \mathbf{A}_t^{e,\tau}
    = (1-\tau)\boldsymbol{\epsilon}
    + \tau\mathbf{A}_t^e.
    \label{eq:flow_interpolation}
\end{equation}
The action Transformer predicts a conditional velocity field
\(v_\theta\), trained with
\begin{equation}
    \mathcal{L}_{\mathrm{FM}}
    =
    \mathbb{E}
    \left[
    \left\|
    v_\theta(\mathbf{A}_t^{e,\tau},\tau,Z_t^e)
    -
    (\mathbf{A}_t^e-\boldsymbol{\epsilon})
    \right\|_2^2
    \right].
    \label{eq:flow_matching}
\end{equation}
Training uses balanced human and robot mini-batches, with the loss
computed over valid action dimensions and non-padded steps.
At inference, the robot policy generates action chunks by integrating
the learned velocity field from Gaussian noise over
\(\tau\in[0,1]\).

Tactile trajectories \(\mathbf{C}_t^e\) from recorded
demonstration segments are used only during training to measure
interaction consistency and guide latent alignment
(Sec.~\ref{sec:action_touch_alignment}).
During deployment, the policy relies only on current observations,
including current tactile measurements, and does not require
demonstration trajectories, DTW, or OT.

\begin{figure}[t]
    \centering
    \includegraphics[width=\columnwidth]{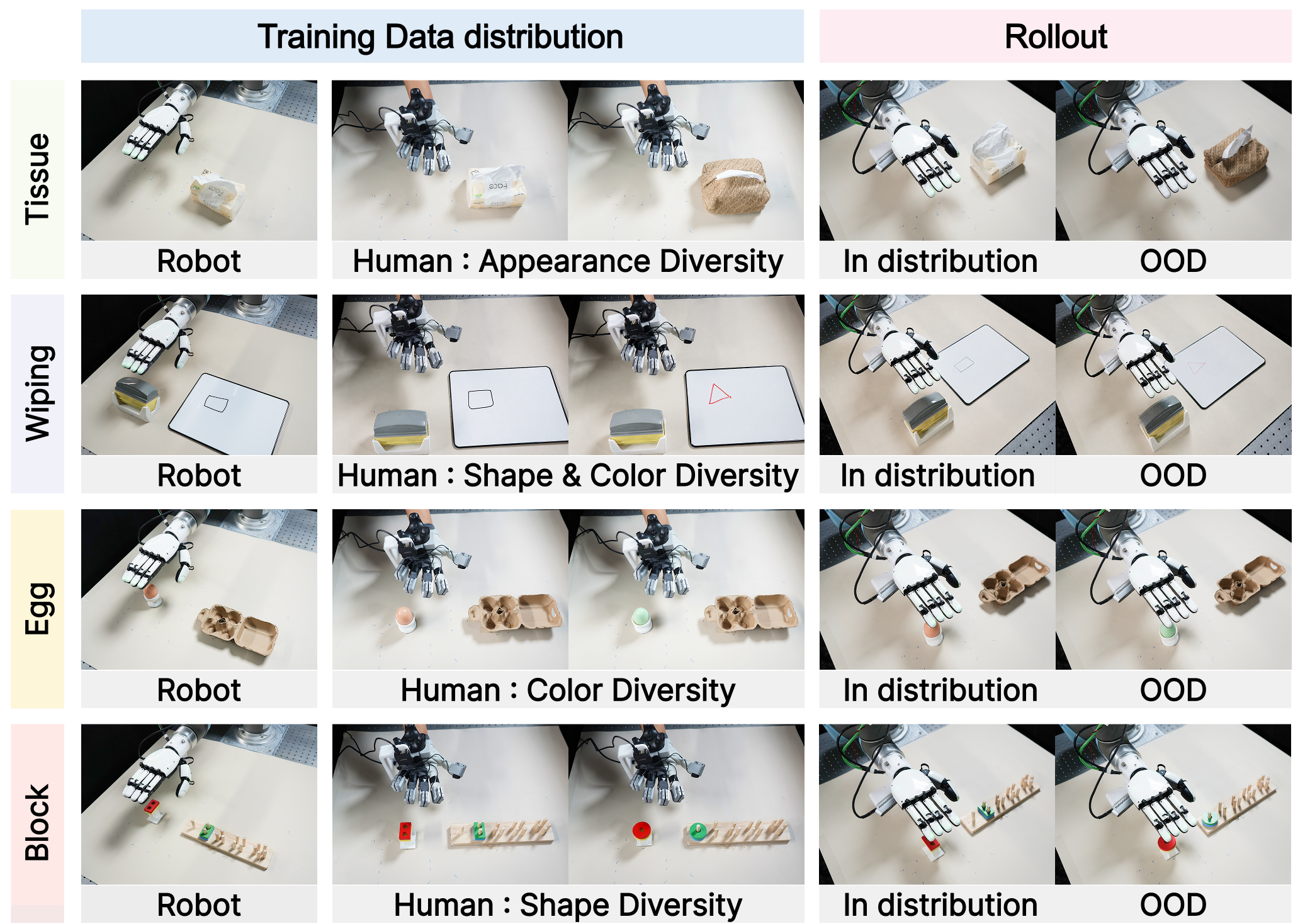}
    \caption{Training distributions and ID/OOD evaluation configurations for
    Tissue, Wiping, Egg, and Block. OOD is defined relative to the robot
    training domain: these configurations are represented in human
    demonstrations but absent from robot training data.}
    \label{fig:data-distribution}
\end{figure}

\subsection{Action--Tactile-Guided Latent Alignment}
\label{sec:action_touch_alignment}

Although human and robot demonstrations are jointly used for policy
learning, their latent representations are not explicitly constrained
to preserve interaction-consistent correspondence. Moreover, similar
motions can involve different contact states and interaction outcomes.
TacOT therefore leverages action and tactile trajectories to identify
interaction-consistent correspondences and guide human--robot latent
alignment.

\textbf{Action--tactile correspondence.}
Within each training mini-batch, we compute action and tactile
DTW distances \(D^a_{ij}\) and \(D^c_{ij}\) for every
human--robot chunk pair. Local costs use weighted action MSE
and tactile MSE; tactile channels are standardized using fixed
per-embodiment training-set statistics. DTW connects valid
sequence endpoints using horizontal, vertical, and diagonal steps
without a warping band; its accumulated cost is divided by path length.
Each distance matrix is robustly normalized using its median and
interquartile range before fusion:
\begin{equation}
    S_{ij}
    =
    \frac{w_a\bar D^a_{ij}+w_c\bar D^c_{ij}}
    {w_a+w_c},
    \qquad
    i^\star(j)=\arg\min_i S_{ij},
\label{eq:action_tactile_matching}
\end{equation}
Here, \(w_a\) and \(w_c\) balance action and tactile similarity.
To anchor correspondence to the target embodiment, each robot chunk
\(j\) identifies one preferred human chunk \(i^\star(j)\) from the mini-batch.
This selection only discounts its latent transport cost in
Eq.~\eqref{eq:dtw_shaped_cost}, preserving soft OT coupling across all pairs.

\textbf{Correspondence-guided optimal transport.}
We first flatten and L2-normalize the encoded context tokens:
\begin{equation}
    z_i^e
    =
    \frac{\operatorname{vec}(Z_i^e)}
    {\|\operatorname{vec}(Z_i^e)\|_2},
    \qquad
    D^z_{ij}
    =
    \|z_i^{\mathrm H}-z_j^{\mathrm R}\|_2^2 ,
\label{eq:latent_distance}
\end{equation}
where \(D^z_{ij}\) denotes the latent transport cost between human and
robot chunks.

To encourage interaction-consistent latent alignment, we reduce
the transport costs of the selected human--robot chunk pairs:
\begin{equation}
    \widetilde C_{ij}
    =
    \begin{cases}
        \lambda D^z_{ij}, & i=i^\star(j),\\
        D^z_{ij}, & \text{otherwise},
    \end{cases}
    \qquad 0<\lambda<1.
    \label{eq:dtw_shaped_cost}
\end{equation}
This cost shaping favors the selected pairs without excluding
other matches, preserving soft many-to-many correspondence.

Let \(\mu_{\mathrm H}\) and \(\mu_{\mathrm R}\) denote uniform
empirical measures over human and robot latent representations,
respectively. We define the alignment loss through
entropy-regularized optimal transport:
\begin{equation}
    \mathcal{L}_{\mathrm{AT\text{-}OT}}
    =
    \operatorname{Sinkhorn}_{\varepsilon}
    (\mu_{\mathrm H},\mu_{\mathrm R};\widetilde C),
    \label{eq:action_touch_ot}
\end{equation}
where \(\varepsilon\) controls entropy regularization and
\(\widetilde C\) incorporates action--tactile correspondence
into the latent transport cost.
We implement this objective in GeomLoss using the customized
cost function~\cite{feydy2019interpolating}.

The joint training objective is
\begin{equation}
    \mathcal{L}_{\mathrm{total}}
    =
    \mathcal{L}_{\mathrm{FM}}
    +
    \alpha\mathcal{L}_{\mathrm{AT\text{-}OT}},
    \label{eq:total_training_loss}
\end{equation}
where \(\alpha\) weights the alignment loss.
Both losses update the observation encoder, while only
\(\mathcal{L}_{\mathrm{FM}}\) updates the action decoder.
DTW and OT are used exclusively during training and
add no inference overhead.

%%%%%%%%%%%%%%%%%%%%%%%%%%%%%%%%%%%%%%%%%%%%%%%%%%%%%%%%%%%%%%%%%%%%%%%%%%%%%%%%

\begin{figure*}[t]
    \centering
    \includegraphics[width=0.95\textwidth]{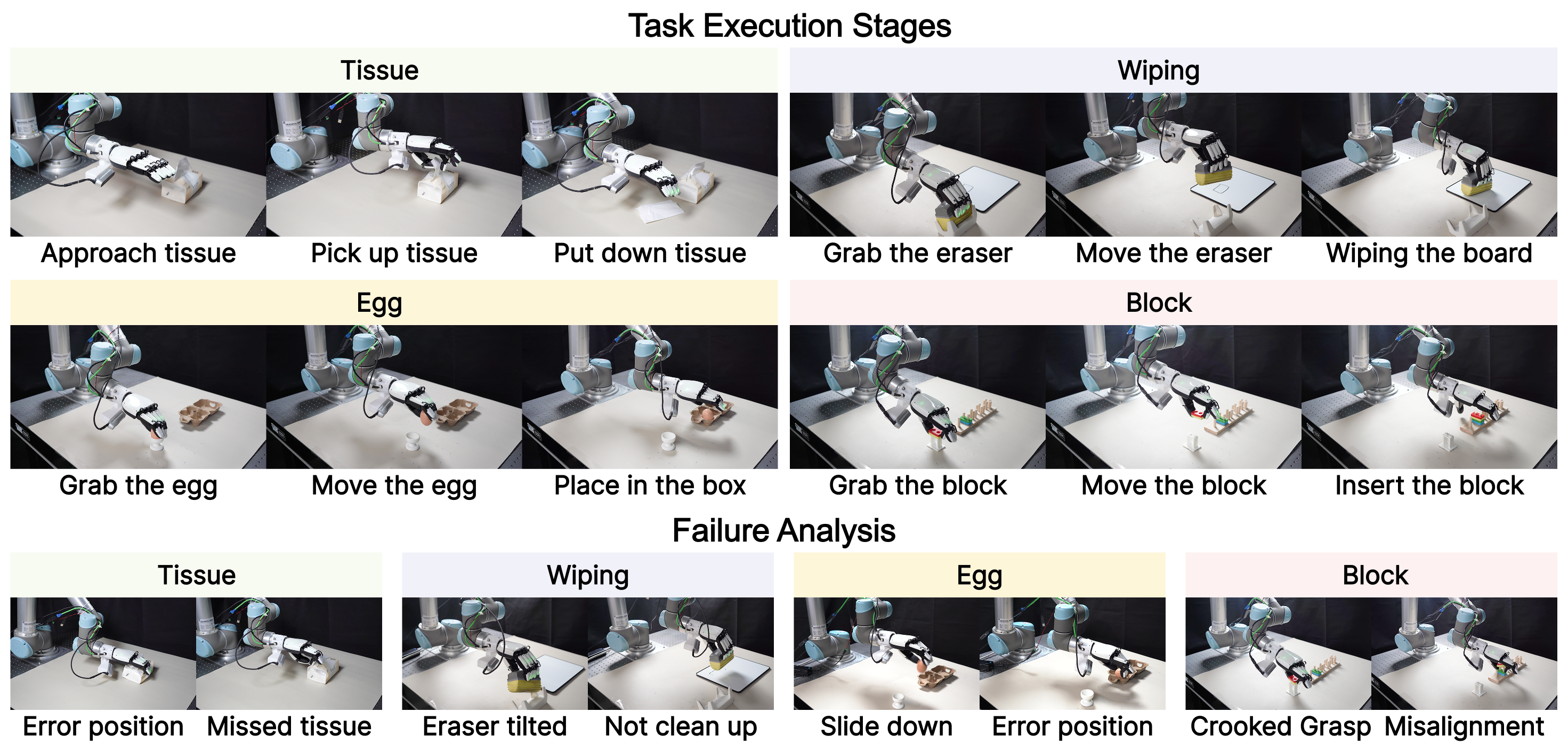}
    \caption{
    Representative real-robot executions across four tasks.
    Top: successful rollouts showing key execution stages for Tissue, Wiping,
    Egg, and Block. Bottom: representative failure cases illustrating
    task-specific failure modes.
    }
    \label{fig:task_sequences}
\end{figure*}

\section{EXPERIMENTS}

We evaluate TacOT on four real-world dexterous manipulation tasks using
closed-loop success rates as the primary metric. The evaluation includes
in-distribution comparisons with representative baselines, targeted
human-to-robot transfer to configurations unseen during robot training
but represented in human demonstrations, analyses of action--tactile
correspondence and latent representations, and experiments with varying
numbers of human and robot demonstrations.

\subsection{Experimental Setup}

\textbf{Tasks and success criteria.}
We evaluate TacOT on four real-world contact-rich manipulation tasks:
\textit{Tissue}, \textit{Wiping}, \textit{Egg}, and \textit{Block}.
These tasks cover diverse contact-rich manipulation challenges, including
flexible-object handling, sustained surface interaction, fragile-object
manipulation, and geometry-sensitive insertion.
All methods use the same UR5--XHand platform and closed-loop control protocol.

We use task-specific binary success criteria. A trial is successful if
the tissue is placed in the designated region, the target marks are
removed during wiping, the egg is placed intact inside the container,
or the block is fully inserted into the corresponding slot.
The same criteria are applied to all methods under both
in-distribution (ID) and out-of-distribution (OOD) conditions.

\textbf{Data splits and evaluation protocol.}
We construct episode-level training and evaluation splits to measure
both ID performance and human-to-robot transfer.
For each task, all human-augmented methods use the same 60 robot ID episodes
and 200 human demonstration episodes, including 100 ID and 100 OOD episodes.
Robot-only uses the same 60 robot ID episodes without human demonstrations.
The methods differ only in their training objectives, correspondence mechanisms,
and tactile observation inputs, as indicated in Table~\ref{tab:main_results}.
OOD is defined relative to the robot training domain: each task uses one
configuration represented in the human demonstrations but excluded from robot
training. We collect 100 human demonstrations for this configuration and
evaluate each method with 15 robot trials under the same setting; the resulting
success rates are reported in Table~\ref{tab:ood_results}. The OOD changes are
different tissue-package appearances for Tissue, a black square changed to a
red triangle for Wiping, an orange egg changed to cyan for Egg, and a
rectangular block changed to circular for Block.

All methods are evaluated in the same scenes and initial poses under
closed-loop control.
We perform 30 rollouts per method--task pair for ID evaluation and
15 rollouts per pair for OOD evaluation.
Control runs at \(30~\mathrm{Hz}\), and complete-task binary success
rate is used as the primary metric.

\begin{figure*}[t]
    \centering
    \includegraphics[width=0.95\textwidth]{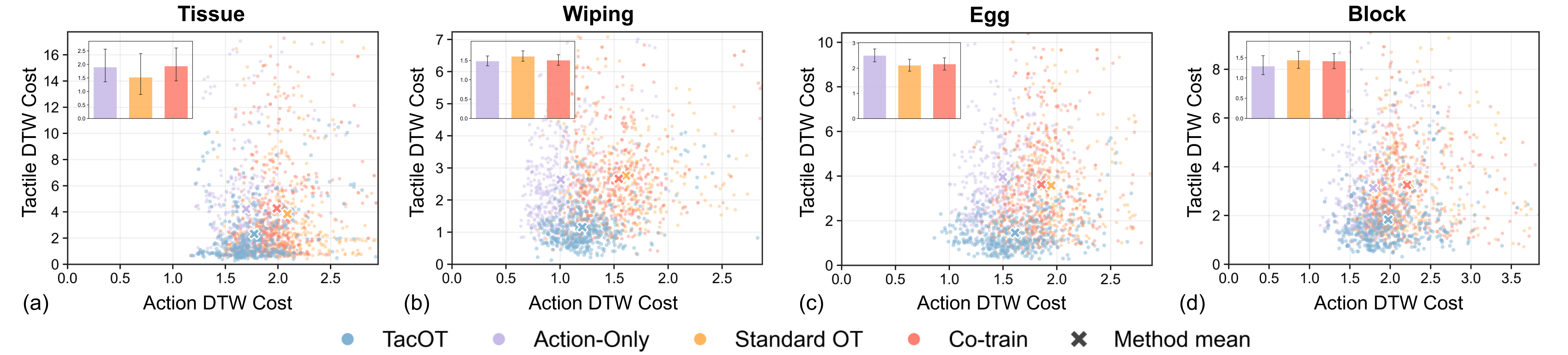}
    \caption{
    Normalized action and tactile mismatch of retrieved human--robot pairs across
    four tasks. ``X'' markers denote method means. Insets show relative tactile mismatch reduction achieved by TacOT.
}
    \label{fig:action-tactile-alignment}
\end{figure*}

% --- TABLE I: MAIN RESULTS ---
% \begin{table}[!t]
%     \centering
%     \caption{In-Distribution Real-World Evaluation Results}
%     \label{tab:main_results}
%     \setlength{\tabcolsep}{4.0pt}
%     \footnotesize
%     \renewcommand{\arraystretch}{1.18}
%     \begin{tabular*}{\columnwidth}{@{\extracolsep{\fill}}lccccc@{}}
%         \toprule
%         \textbf{Method} & \textbf{Tactile Obs.} & \textbf{Tissue} & \textbf{Wiping} & \textbf{Egg}
%         & \textbf{Block} \\
%         \midrule
%         Robot-only                  & \cmark & 60\% & 53\% & 47\% & 47\% \\
%         Co-train                    & \cmark & 57\% & 53\% & 37\% & 40\% \\
%         Standard OT                 & \cmark & 83\% & 70\% & 53\% & 57\% \\
%         MimicPlay~\cite{mimicplay} & \xmark & 47\% & 50\% & 23\% & 27\% \\
%         EgoMimic~\cite{egomimic} & \xmark & 67\% & 60\% & 37\% & 43\% \\
%         Action-guided OT                 & \xmark & 80\% & 77\% & 43\% & 50\% \\
%         Action-guided OT                 & \cmark & 87\% & \textbf{83\%} & 60\% & 63\% \\
%         \textbf{TacOT}              & \textbf{\cmark} & \textbf{93\%} & \textbf{83\%} &
%                                       \textbf{73\%} & \textbf{80\%} \\
%         \bottomrule
%     \end{tabular*}
% \end{table}

\begin{table}[!t]
    \centering
    \caption{In-Distribution Real-World Evaluation Results}
    \label{tab:main_results}
    \setlength{\tabcolsep}{4.0pt}
    \footnotesize
    \renewcommand{\arraystretch}{1.18}
    \begin{tabular*}{\columnwidth}{@{\extracolsep{\fill}}lcccc@{}}
        \toprule
        \textbf{Method} & \textbf{Tissue} & \textbf{Wiping} & \textbf{Egg}
        & \textbf{Block} \\
        \midrule
        Robot-only & 60\% & 53\% & 47\% & 47\% \\
        Co-train & 57\% & 53\% & 37\% & 40\% \\
        Standard OT & 83\% & 70\% & 53\% & 57\% \\
        MimicPlay~\cite{mimicplay} & 47\% & 50\% & 23\% & 27\% \\
        EgoMimic~\cite{egomimic} & 67\% & 60\% & 37\% & 43\% \\
        AOT (w/o tactile obs.) & 80\% & 77\% & 43\% & 50\% \\
        AOT (w/ tactile obs.) & 87\% & 83\% & 60\% & 63\% \\
        \textbf{TacOT} & \textbf{93\%} & \textbf{83\%} &
        \textbf{73\%} & \textbf{80\%} \\
        \bottomrule
    \end{tabular*}
\end{table}

\textbf{Implementation details.}
All variants use the same episode-level data splits and optimization
protocol. Tactile-observed variants use the multimodal observation space.
MimicPlay, EgoMimic, and the action-guided OT variant without tactile
observations exclude tactile inputs. The policy
uses a 16-layer
Transformer observation encoder and an 8-layer Transformer action
decoder, both with hidden dimension 256 and 8 attention heads.
For TacOT, we set $w_a=0.7$, $w_c=0.3$, $\alpha=0.7$, and $\lambda=0.1$.
During deployment, the controller runs at 30~Hz and executes the first
8 actions of each predicted chunk before replanning. DTW and OT are
used only during training and add no inference-time computation.

\textbf{Compared methods.}
We compare TacOT with seven baselines.
Robot-only learns solely from robot demonstrations, whereas
Co-train jointly trains on human and robot demonstrations without
explicit human--robot alignment.
Standard OT augments joint training with latent Sinkhorn
optimal transport without correspondence guidance.
Robot-only, Co-train, and Standard OT use current tactile
observations during policy learning.
MimicPlay~\cite{mimicplay} and EgoMimic~\cite{egomimic}
provide additional human-to-robot imitation baselines; neither
uses tactile observations in our evaluation.
Action-guided OT (AOT, w/ tactile obs.) is our primary baseline.
It shares TacOT's policy architecture, current tactile
observations, and OT alignment objective, but uses action-only DTW correspondence without tactile trajectory matching.
AOT (w/o tactile obs.) further removes tactile policy inputs and
serves as an EgoBridge-style baseline~\cite{egobridge}.
The two AOT variants assess the effect of tactile policy inputs,
while comparing TacOT with AOT (w/ tactile obs.) assesses the
contribution of tactile-guided correspondence.

\subsection{Real-Robot Manipulation Evaluation}
% This subsection answers Q1 and Q2 using ID and targeted OOD rollouts.

\subsubsection{In-Distribution Performance}

Table~\ref{tab:main_results} reports closed-loop success rates
under the in-distribution (ID) protocol. TacOT achieves the
best or tied-best performance across all four tasks:
93\% on \textit{Tissue}, 83\% on \textit{Wiping}, 73\% on
\textit{Egg}, and 80\% on \textit{Block}. Standard OT
outperforms Co-train on every task, suggesting that explicit
representation alignment can improve human--robot transfer
compared with joint training alone in this evaluation.

MimicPlay and EgoMimic provide competitive human-to-robot
imitation baselines without tactile observations. Compared with
these methods, TacOT achieves higher success rates across all
tasks, particularly on contact-sensitive tasks such as
\textit{Egg} and \textit{Block}, where tactile dynamics provide
additional interaction-aware cues for correspondence learning.

To assess the contribution of tactile-guided correspondence,
we compare TacOT with AOT (w/ tactile obs.). Both methods use
the same policy architecture, tactile observations, and OT
alignment objective, differing only in whether tactile trajectories
over the prediction horizon guide correspondence selection during training.
TacOT improves over AOT on \textit{Tissue}, \textit{Egg}, and
\textit{Block}, while achieving the same success rate on
\textit{Wiping}. The largest gains occur on \textit{Egg} and
\textit{Block}, with improvements of approximately 13 and 17
percentage points, respectively. These results support the
contribution of tactile-guided correspondence beyond tactile
policy conditioning alone.

We separately evaluate the role of tactile policy inputs by
comparing AOT (w/o tactile obs.) and AOT (w/ tactile obs.).
Adding tactile inputs improves success across all four tasks,
with the largest gains on \textit{Egg} (43\% to 60\%) and
\textit{Block} (50\% to 63\%). This comparison suggests that
tactile observations benefit closed-loop execution, complementing
their role in correspondence learning. Representative successful
executions and task-specific failure cases are shown in
Fig.~\ref{fig:task_sequences}.

\begin{table}[t]
    \centering
    \caption{Human-to-Robot OOD Evaluation Results}
    \label{tab:ood_results}
    \setlength{\tabcolsep}{4.0pt}
    \footnotesize
    \renewcommand{\arraystretch}{1.18}
    \begin{tabular*}{\columnwidth}{@{\extracolsep{\fill}}lcccc@{}}
        \toprule
        \textbf{Method} & \textbf{Tissue} & \textbf{Wiping} &
        \textbf{Egg} & \textbf{Block} \\
        \midrule
        Robot-only     & 27\% & 13\% & 0\% & 0\% \\
        Action-guided OT    & 60\% & 47\% & 27\% & 33\% \\
        \textbf{TacOT} & \textbf{73\%} & \textbf{60\%} &
        \textbf{40\%} & \textbf{53\%} \\
        \bottomrule
    \end{tabular*}
\end{table}

\subsubsection{Human-to-Robot Transfer}

To evaluate targeted human-to-robot transfer, we evaluate the trained
policies on configurations represented in human demonstrations but absent
from robot training (Fig.~\ref{fig:data-distribution}). We compare Robot-only,
action-guided OT, and TacOT, all with tactile policy observations,
using the same success criteria as in the ID evaluation; results are reported in Table~\ref{tab:ood_results}. This
protocol evaluates transfer to robot-unseen configurations.

\subsection{Action--Tactile Alignment Analysis}
\label{sec:mechanism-analysis}

\subsubsection{Offline Correspondence Analysis}

We compare retrieved human--robot pairs using identical validation Robot queries
and Human candidate pools. For each task, we retain Robot chunks above the
task-specific 75th-percentile contact-intensity threshold and sample 2,000 Human
validation chunks per query. For this correspondence analysis, action-guided OT retrieves the
candidate with minimum action DTW distance; Standard OT and Co-train
retrieve latent nearest neighbors, and TacOT retrieves the candidate
with minimum action--tactile matching score. We evaluate all pairs using normalized action and
tactile mismatch, where lower values indicate greater similarity.

As shown in Fig.~\ref{fig:action-tactile-alignment}, TacOT selected pairs with
lower tactile DTW mismatch across all four tasks while maintaining a comparable
action-mismatch range. Relative to action-guided OT, TacOT reduced the mean tactile DTW
mismatch from 4.227 to 2.325 on Tissue, from 2.644 to 1.158 on Wiping, from 3.963
to 1.457 on Egg, and from 3.129 to 1.835 on Block, corresponding to relative
reductions of 45.0\%, 56.1\%, 63.2\%, and 41.3\%, respectively. TacOT also
achieved lower mean action and tactile DTW mismatch than Standard OT and Co-train.
These results indicate that tactile-aware pairing improves contact similarity
without substantially compromising action similarity.

For each task and method, we randomly sample 420 human--robot chunk pairs;
colored crosses denote sample means. Error bars show 95\% confidence intervals
from robot-episode-level clustered bootstrap resampling.

\begin{figure}[t]
    \centering
    \includegraphics[width=\columnwidth]{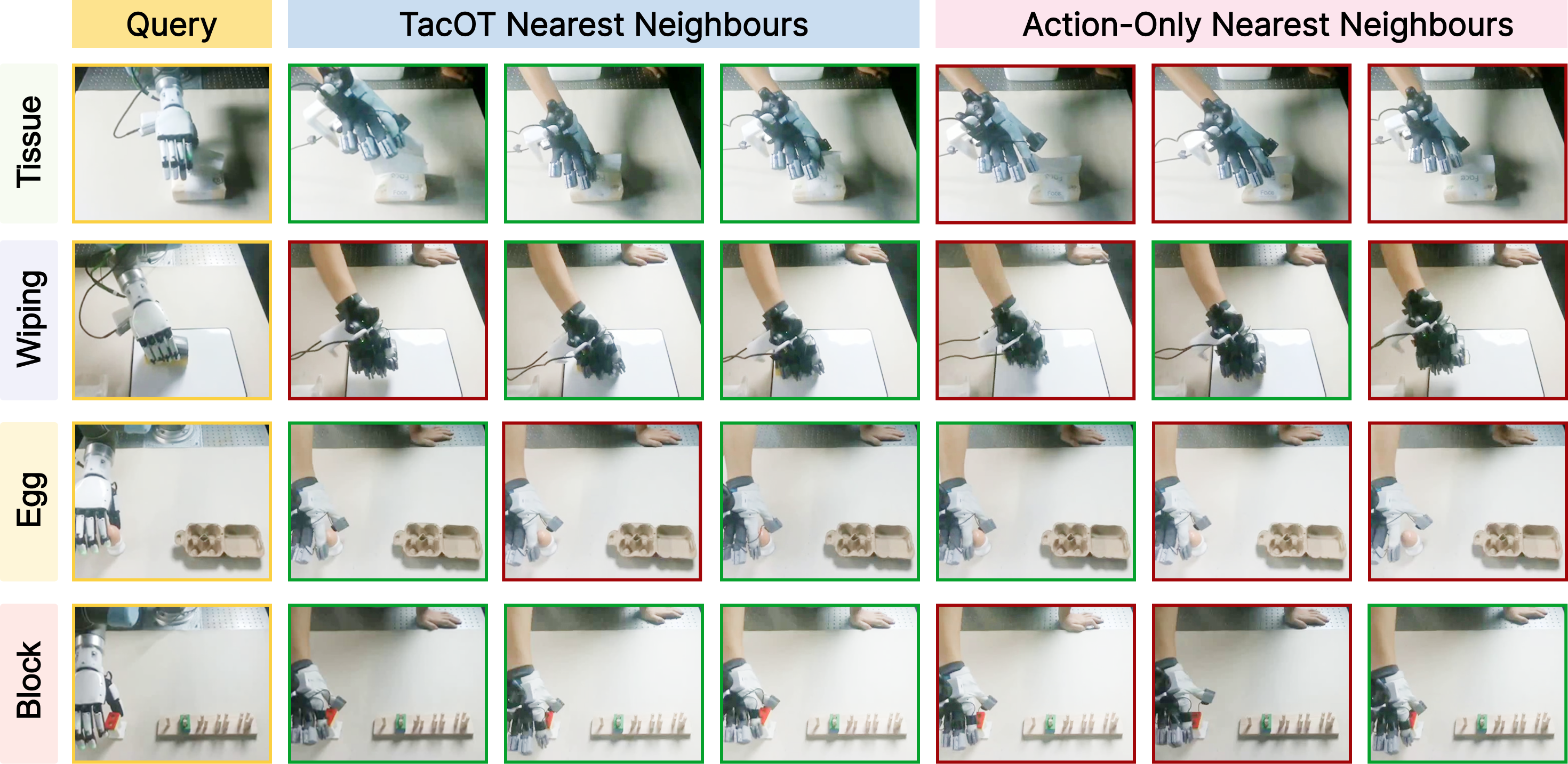}
    \caption{
    Nearest-neighbour retrievals from the learned latent spaces.
    Green and red borders indicate interaction-consistent and
    interaction-inconsistent pairs, respectively.
    }
    \label{fig:nearest-neighbour-retrieval}
\end{figure}

\subsubsection{Offline Model Analysis}

After training, we analyze whether the latent space learned by TacOT
captures behavioral structure jointly determined by motion patterns and
interaction dynamics. Figure~\ref{fig:nearest-neighbour-retrieval} shows
nearest-neighbour retrievals from the learned latent space across the four
tasks. Green and red borders indicate interaction-consistent and
interaction-inconsistent retrievals, respectively. TacOT retrieves more
interaction-consistent human neighbours than action-guided OT across all tasks.
Tactile cues help distinguish demonstrations with similar motions but
different interaction states, such as approaching versus established
contact in Wiping and grasping versus placing in Egg.

To quantitatively evaluate whether latent-space neighbours preserve interaction dynamics, we compare latent-space distances with contact-evolution
distances. For each robot query window $q$ and human candidate
$h_j$, we compute the latent-space distance
$d^z_{q,j}=\lVert z_q^R-z_j^H\rVert_2^2$, where $z_q^R$ and $z_j^H$
are the corresponding latent representations.
We extract the following $H=12$ time steps of fingertip force/contact
trajectories, denoted by $f_q^R$ and $f_j^H$, and compute
the contact-dynamics distance $d^f_{q,j}=\lVert f_q^R-f_j^H\rVert_2^2$.
After filtering invalid timesteps and normalization, this distance measures
contact-evolution differences independently of force magnitude.

\begin{figure}[t]
    \centering
    \includegraphics[width=0.98\columnwidth]{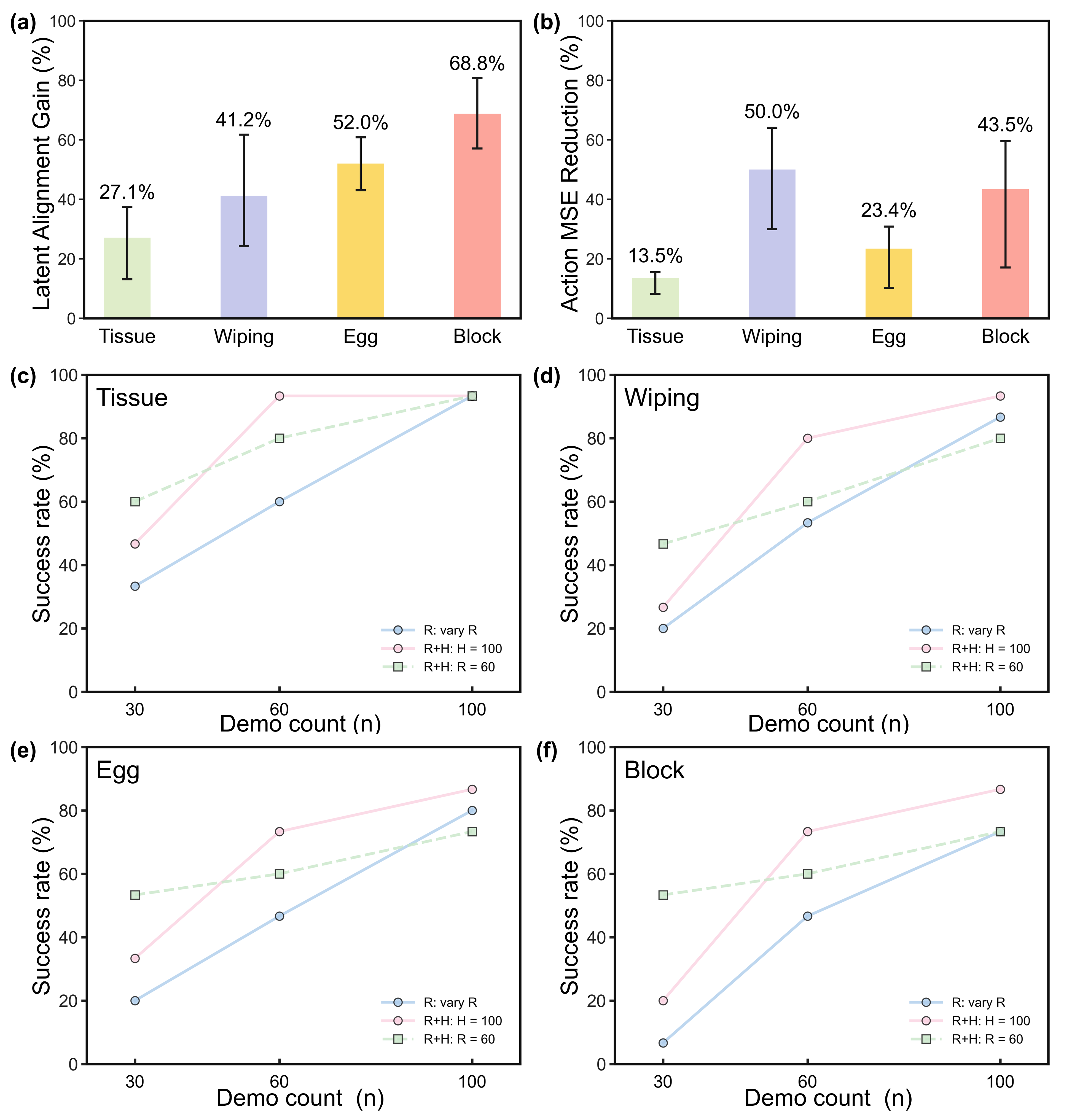}
    \caption{
    Quantitative analysis of latent alignment, replanning consistency,
    and demonstration efficiency. (a) Latent--contact-evolution alignment,
    (b) overlap action-plan MSE reduction, and (c)--(f) success rates
    under varying numbers of human and robot demonstrations for Tissue,
    Wiping, Egg, and Block, respectively.
    }
    \label{fig:quantitative-analysis}
\end{figure}

For each query, candidates are ranked by $d^z_{q,j}$ and $d^f_{q,j}$,
yielding rankings $r_q^z$ and $r_q^f$. We measure their consistency using
the Spearman coefficient $\rho_q=\operatorname{Spearman}(r_q^z,r_q^f)$,
where a higher value indicates better agreement between latent
neighbourhoods and neighbourhoods defined by interaction dynamics. For each task, we sample 2,000 robot
query windows with identical human candidate sets for TacOT and action-guided OT,
aggregate $\rho_q$, and report the relative improvement with 95\% confidence
intervals estimated by a robot-episode-level clustered bootstrap.
As shown in Fig.~\ref{fig:quantitative-analysis}(a), TacOT improves
latent--contact-evolution rank alignment on all four tasks, with relative gains of
27.1\%, 41.2\%, 52.0\%, and 68.8\% on Tissue, Wiping, Egg, and Block,
respectively. These results indicate that tactile dynamics improve the
ability of the latent space to preserve interaction dynamics.

Finally, we evaluate whether the learned latent organization improves
consistency during successive replanning. For two temporally adjacent query
windows $q$ and $q'$, we compare their generated hand-action sequences over
the shared prediction horizon and compute the overlap action-plan MSE.
Lower MSE indicates more consistent action prediction. As shown in
Fig.~\ref{fig:quantitative-analysis}(b), TacOT reduces the overlap
action-plan MSE by 13.5\%, 50.0\%, 23.4\%, and 43.5\% on Tissue, Wiping,
Egg, and Block, respectively.

\subsection{Demonstration Efficiency Analysis}

The demonstration-efficiency results in Fig.~\ref{fig:quantitative-analysis}(c)--(f)
use only ID demonstrations. We vary the number of robot demonstrations with
100 human demonstrations fixed, and vary the number of human demonstrations
with 60 robot demonstrations fixed. Increasing robot demonstrations consistently
improves performance across all tasks. Additional human demonstrations are most
beneficial when robot data is limited: with 60 robot demonstrations, adding human
data raises Block success from 47\% to 73\%. These results suggest that robot
demonstrations provide stronger task-specific execution supervision, while human
demonstrations provide complementary behavioral information.

When robot demonstrations are fixed, increasing human demonstrations also
improves joint-training performance. For example, increasing human demonstrations
from 30 to 100 raises success from 47\% to 80\% on Wiping and from 53\% to 73\%
on Block. However, with more robot demonstrations, the marginal benefit of
additional human data becomes smaller.
                                                   
%%%%%%%%%%%%%%%%%%%%%%%%%%%%%%%%%%%%%%%%%%%%%%%%%%%%%%%%%%%%%%%%%%%%%%%%%%%%%%%%
\section{CONCLUSION}

We presented \textbf{TacOT}, a framework that uses action--tactile
correspondence to guide optimal transport alignment for learning
dexterous manipulation from human demonstrations.
Across four real-world tasks, TacOT achieves the best or tied-best
in-distribution success rates among the evaluated methods and
improves transfer to human-demonstrated configurations unseen in
robot training. Compared with action-guided OT, TacOT incorporates tactile dynamics
into correspondence selection, yielding more contact-consistent matches,
stronger alignment between latent similarity and interaction contact dynamics, and more consistent action predictions across successive replanning steps.
Together, these results support the use of tactile information
as correspondence supervision for human-to-robot policy learning.

The current evaluation focuses on human-to-robot transfer with a 
fixed robot embodiment and relies on RGB observations without explicit 3D geometry. Future work will extend TacOT to heterogeneous robot morphologies and incorporate depth and multi-view geometric representations to improve correspondence robustness under more general human-to-robot transfer settings.

%%%%%%%%%%%%%%%%%%%%%%%%%%%%%%%%%%%%%%%%%%%%%%%%%%%%%%%%%%%%%%%%%%%%%%%%%%%%%%%%

\bibliographystyle{IEEEtran}
\bibliography{references}

\end{document}